\documentclass[conference]{IEEEtran}
\IEEEoverridecommandlockouts
\usepackage{cite}
\usepackage{amsmath,amssymb,amsfonts}
\usepackage{algorithmic}
\usepackage{graphicx}
\usepackage{textcomp}
\usepackage{xcolor}

\usepackage{graphicx}
\usepackage{comment}
\usepackage{amsmath,amssymb} 
\usepackage{color}

\usepackage{amssymb}
\usepackage{pifont}
\newcommand{\cmark}{\ding{51}}%
\newcommand{\xmark}{\ding{55}}%

\def\BibTeX{{\rm B\kern-.05em{\sc i\kern-.025em b}\kern-.08em
    T\kern-.1667em\lower.7ex\hbox{E}\kern-.125emX}}
\begin{document}

\title{Gait Recognition using Multi-Scale Partial Representation Transformation with Capsules}

\author{\IEEEauthorblockN{Alireza Sepas-Moghaddam\IEEEauthorrefmark{1},
Saeed Ghorbani\IEEEauthorrefmark{2}, Nikolaus F. Troje \IEEEauthorrefmark{2} and
Ali Etemad\IEEEauthorrefmark{1}}
\\
\IEEEauthorblockA{ \IEEEauthorrefmark{1} Department of Electrical and Computer Engineering \& Ingenuity Labs, Queen's University, Kingston, Ontario, Canada\\
\IEEEauthorrefmark{2} Biomotion Lab, Centre for Vision Research, York University, Toronto ON, Canada}
}




\maketitle

\begin{abstract}
Gait recognition, referring to the identification of individuals based on the manner in which they walk, can be very challenging due to the variations in the viewpoint of the camera and the appearance of individuals. Current methods for gait recognition have been dominated by deep learning models, notably those based on partial feature representations. In this context, we propose a novel deep network, learning to transfer multi-scale partial gait representations using capsules to obtain more discriminative gait features. Our network first obtains multi-scale partial representations using a state-of-the-art deep partial feature extractor. It then recurrently learns the correlations and co-occurrences of the patterns among the partial features in forward and backward directions using Bi-directional Gated Recurrent Units (BGRU). Finally, a capsule network is adopted to learn deeper part-whole relationships and assigns more weights to the more relevant features while ignoring the spurious dimensions. That way, we obtain final features that are more robust to both viewing and appearance changes. The performance of our method has been extensively tested on two gait recognition datasets, CASIA-B and OU-MVLP, using four challenging test protocols. The results of our method have been compared to the state-of-the-art gait recognition solutions, showing the superiority of our model, notably when facing challenging viewing and carrying conditions.
\end{abstract}

\begin{IEEEkeywords}
Gait Recognition, Convolutional Networks, Gated Recurrent Units, Capsule  Network
\end{IEEEkeywords}

\section{Introduction}
Gait data, acquired from an individual's body movement during walking, can provide important identity information that enables a wide range of possibilities for biometric recognition \cite{R4, nambiar}. Gait data can generally be acquired using wearable or non-wearable sensors \cite{R2}. Gait recognition systems based on wearable sensors need cooperation from the users to attach sensors to their body \cite{wear}, while non-wearable gait recognition systems such as that considered in this paper, can acquire gait data with no cooperation from individuals, notably using one or multiple imaging sensors \cite{B6}. There are two challenging factors that can affect the performance of the non-wearable image-based gait recognition systems namely variations in the viewpoint of the camera and the appearance of the individual \cite{R2, ICPR1}. 
Due to the unconstrained nature of gait recognition \cite{TanmayThesis}, gait data can be captured from different viewpoints, so some parts of the body can be hidden from one view to another \cite{B6,view}.
The appearance of individuals can also be different due to variations in clothing, for instance wearing a coat or hat, or carrying a handbag or backpack \cite{R6}.

\begin{figure}[!t]
    \begin{center}
    \includegraphics[width=1\linewidth]{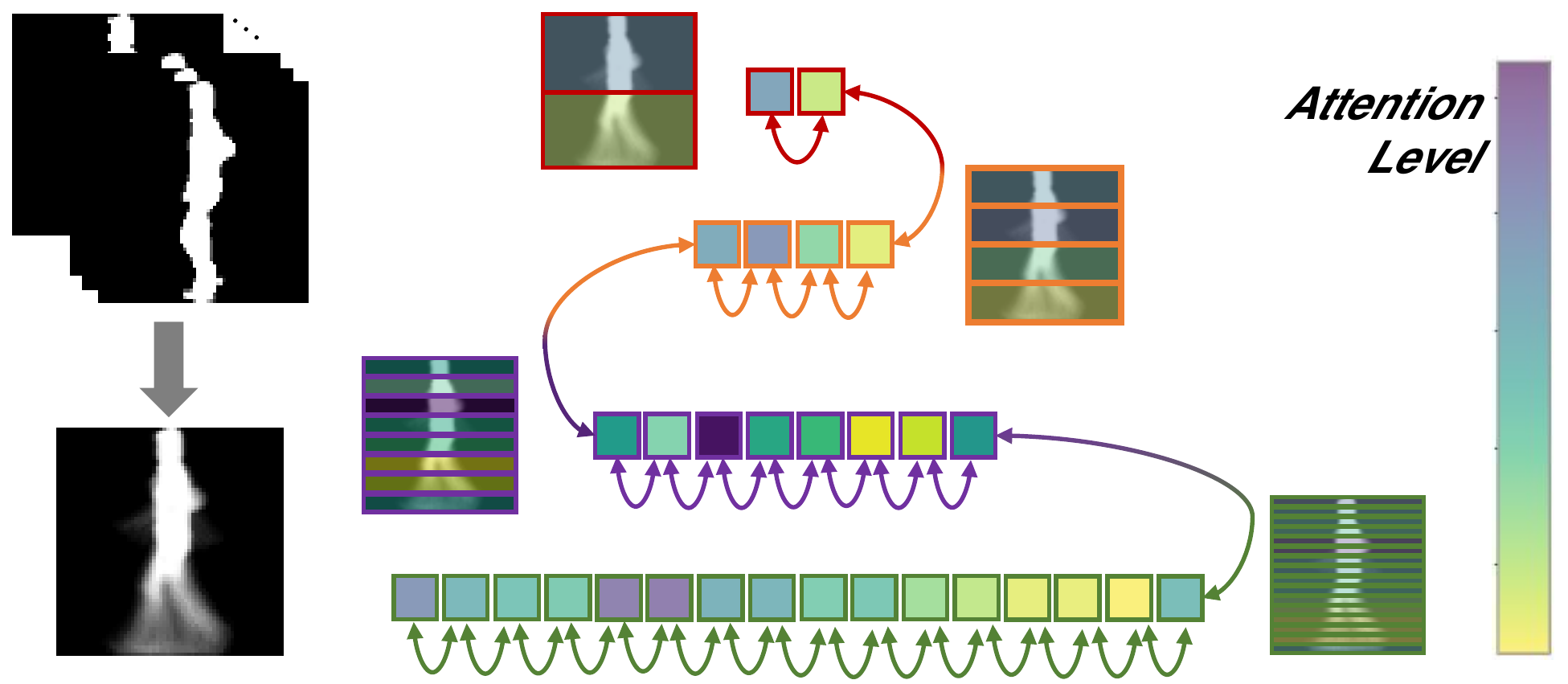} 
    \end{center}
\caption{Overview of our proposed method. The gait silhouette are processed by GitSet method \cite{B1}, extracting partial embeddings at different scales. The relations between partial embedding has then been learned using a BGRU \cite{GRU1}. Finally, the recurrently learned features are further learned by capsules to exploit part-whole relationships and to assign more weights to the more relevant features while ignoring the spurious dimensions.}
\label{fig:1}
\end{figure}

\begin{table*}
  \centering
  \setlength\tabcolsep{6pt}
    \caption{Overview of state-of-the-art deep gait recognition methods.}
    \begin{tabular}{l|l|l|l|l|l}
    \hline
    \textbf {Method}& \textbf{Year}& \textbf{Input}& \textbf{Representation} & \textbf{Network Architecture} & \textbf{Temporal Template} \\
    \hline\hline
    GEINet~\cite{B8} & 2016 & Silhouette & Global &   2 Conv.+2 Pool.+2 FC& GEI \\
    \hline
    Ensemble CNNs~\cite{B6} & 2017 & Silhouette & {\vtop{\hbox{\strut Global}\hbox{\strut Partial}}}   &  {\vtop{\hbox{\strut 3 Conv.+2 Pool.+2 FC}\hbox{\strut 2 Conv.+2 Pool.+1 FC}}}& {\vtop{\hbox{\strut GEI}\hbox{\strut CGIs}}} \\
    \hline
    DiGGAN~\cite{B9} & 2018& Silhouette & Global  &  Auto-encoder & GEI 
      \\
    \hline
    MGANs~\cite{B7} & 2019& Silhouette & Global &  4 Conv.+1 Pool.+3 FC & PEI  \\
    \hline
    EV-Gait~\cite{B2} & 2019& Silhouette & Global & 3 Conv.+3 Pool.+3 Caps. & GEI  \\
    \hline
    Gait-joint~\cite{B10} & 2019 & Silhouette & Global  &  12 Conv.+5 Pool.+3 FC & Conv. GEM  \\
    \hline
    GaitNet-1~\cite{B3} & 2019& Silhouette; Skeleton & Global &   Auto-encoder+LSTM & -- \\
    \hline
    Gait-Set~\cite{B1} & 2019& Silhouette & Partial &  6 Conv.+2 Pool.+1 FC & Conv. GEM \\
    \hline
    GaitNet-2~\cite{B4} & 2020& Silhouette; Skeleton  & Global &   Auto-encoder+LSTM & -- \\
    \hline
    \end{tabular}%
  \label{tab1}%
\end{table*}%

Gait recognition with deep learning has gained momentum in recent years, providing state-of-the-art recognition accuracy \cite{R4}. However, due to the challenging factors stated above, it is hard to learn gait features that are highly robust to view and appearance changes \cite{B6}. Deep learning methods based on partial feature representations have been proven to be effective for view- and appearance- invariant gait recognition \cite{B6,R9,B1}. These methods are based on splitting gait data into spatial and/or temporal bins, possibly at multiple scales, to obtain partial gait representations. The partial representations are generally concatenated in a single feature vector to perform gait recognition \cite{B6,R9,B1}. 
However, these representations could be further processed to explore their hierarchical part-whole relations, while preserving positional attributes, such as location, rotation, and scale of the representations. By preserving positional attributes, the gait recognition systems could be more robust to the view and orientation changes of the gait data that could be instrumental for designing improved view-invariant gait recognition systems. The importance of partial features could also be investigated as they may represent different amounts of person-specific information which may not contribute equally towards the final recognition performance \cite{part}; this issue is more evident when a feature represents an appearance changing factor, e.g., a backpack.

To address these issues, in this paper, we propose a novel deep network, learning to transfer partial gait representations using capsules to obtain more discriminative gait features robust to both viewing and appearance changes. Figure \ref{fig:1} presents an abstract depiction of how the learned representations are transformed throughout our proposed solution. Our deep network first uses a state-of-the-art deep partial feature extractor, Gaitset \cite{B1}, to obtain multi-scale partial representations. The partial representations then form a sequence of features to feed a bi-directional Gated Recurrent Units (BGRU) \cite{GRU1}, thus recurrently learning the correlations and co-occurrences of the patterns among the partial feature in forward and backward directions for each specific input representation. A capsule network \cite{caps} is then used, treating recurrently learned partial features as capsules to learn coupling weights between the underlying and output capsules through dynamic routing. It develops hierarchical part-whole relationships between the partial representations that makes our network more robust against view changes. Additionally, the capsule network assigns more weights to the more relevant features while ignoring the spurious dimensions, thus obtaining gait representations with more robustness against appearance changes. A softmax classifier is finally used to perform gait recognition. Although the combination of BGRU and capsule networks has recently been applied in a few other areas such as question target classification \cite{GRUCaps1}, sentiment classification \cite{GRUCaps2}, and emotion detection \cite{GRUCaps3}, it has never used in the context of a biometric recognition system.

Our contributions can be summarized as follows:
\begin{enumerate}
    \item For the first time, we use a Recurrent Neural Network (RNN) to learn the relations between multi-scale partial gait representations. We then use a capsule network to learn deeper part-whole relationships and to act as an attention mechanism to make our model more robust to different viewing and appearance conditions;
    \item We apply Uniform Manifold Approximation and Projection (UMAP) visualization to the learned capsule representations of gait for the first time, showing the discriminative power of our proposed solution;
    \item Our model obtains state-of-the-art values for gait recognition on two gait datasets. Our proposed model is robust not only to view-point changes, but also to human appearance changes including wearing different clothing and carrying conditions.
\end{enumerate}

The rest of the paper is organized as follows. In the next section, we describe the related works and recent research on gait recognition. Following, we present our proposed model, followed by a description of the experiments performed to evaluate our method. Then, the results are provided along with a comparison to the state-of-the-art. Ablation experiments are then discussed in detail. Lastly, the summary is presented.

\section{Related Work}\label{sec:related_work}
Gait recognition solutions can be classified into model-based \cite{Model} and appearance-based \cite{B6} categories, respectively dealing with the structure and appearance of the body. Nowadays, deep learning networks provide state-of-the-art results on both model-based and appearance-based gait recognition \cite{R4}. Table \ref{tab1} presents an overview of the main characteristics of deep learning methods for gait recognition, sorted according to their release date, highlighting their network architectures and the templates used to aggregate temporal information along with the video. These temporal templates can be obtained either in the initial layer of the deep network in the form of Gait Energy Image (GEI) \cite{B6,R7, ICPR2}, Period Energy Image (PEI) \cite{B7}, Chrono Gait Image (CGI) \cite{GEI1}, and others, or in an intermediate layer of the network, for instance in the form of Convolutional Gait Energy Map (Conv. GEM). 

In term of representation, gait recognition methods can have either global or partial representations. Gait recognition methods based on global representations deal with gait data as a whole; some examples include MGANs ~\cite{B7}, GaitNet-1 ~\cite{B3}, GaitNet-2~\cite{B4}, and DiGGAN~\cite{B9}. Gait recognition methods based on partial representations, such as Ensemble CNNs~\cite{B6} and Gait-Set ~\cite{B1},  split gait data into spatial (and/or temporal) bins. The split data can then be further processed, for instance using Fully Connected (FC) layers, thus extracting partial features that are less sensitive to pose and appearance variations.

\begin{figure*}[!t]
    \begin{center}
    \includegraphics[width=1\linewidth]{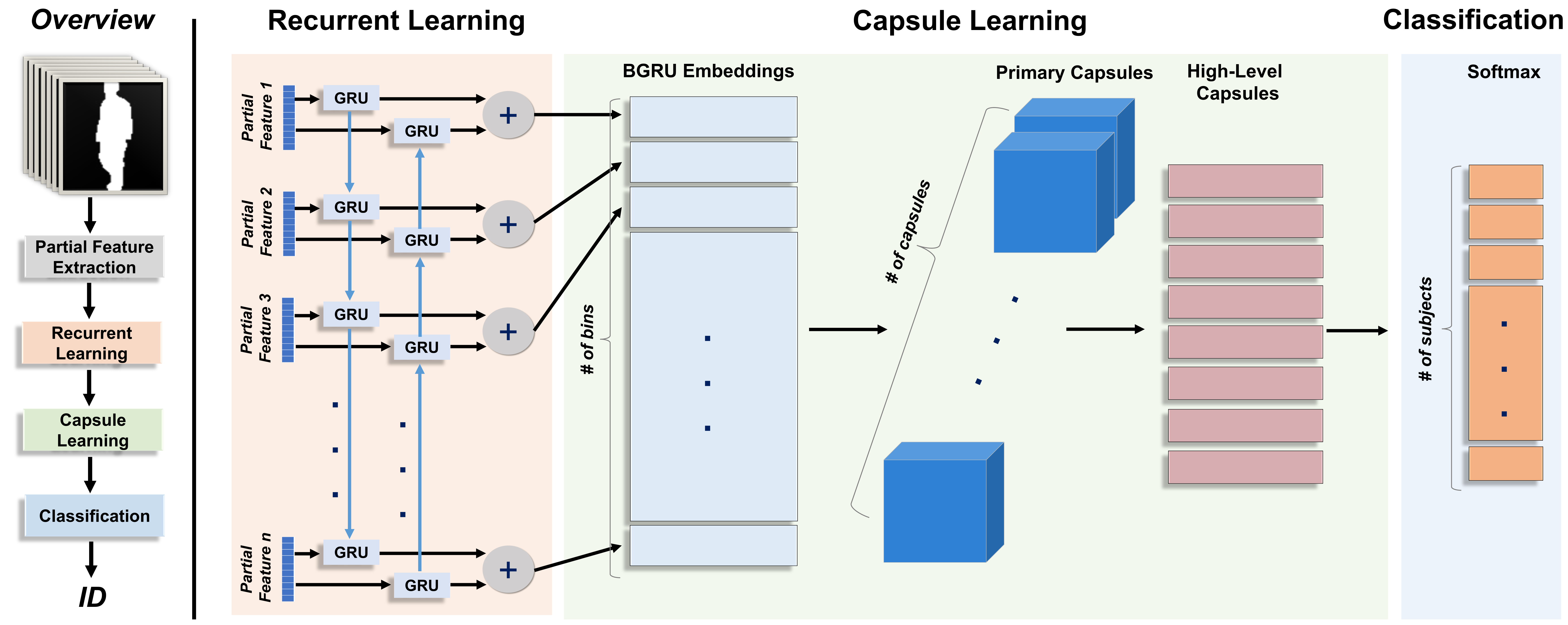} 
    \end{center}
\caption{Architecture of our gait recognition network.}
\label{fig:2}
\end{figure*}

Among the reviewed solutions, GaitSet \cite{B1} currently outperforms other recognition solutions for cross-view gait recognition on two large-scale CASIA-B and OU-MVLP gait datasets. It first extracts convolutional maps from frame-level gait silhouettes and then uses a set-pooling strategy to aggregate convolutional spatio-temporal information in the form of a Conv. GEM template. The template is split into bins at different scales to be then mapped to a discriminative feature space with a horizontal pyramid matching scheme \cite{R9}, forming the final gait representation, where the learned bins' features have the same importance towards the final goal. Our deep network uses Gaitset \cite{B1} as its first block to obtain multi-scale partial representations to be then transferred to a more discriminative feature space using capsules.

\section{Method}\label{sec:method}

\subsection{Problem Definition}\label{sec:problem_definition}
Given a dataset $\mathcal{D}$ of gait sequences where each sequence $S_{i, j}$ is recorded from an identity $x_i \in \{x_1,x_2,\dots, x_N\}$ and a recording view angle $\theta_j \in \{\theta_{1}, \theta_{2}, \dots, \theta_{V}\}$, the task of gait recognition for a probe gait sequence $S_{probe}$ can be defined as:
\begin{equation} \label{eq:task}
    \hat{x} = arg \max\limits_{x_i} Pr(x_i|S_{probe}, \mathcal{D}),
\end{equation}
where $Pr(x_i|S_{probe}, \mathcal{D})$ is the probability of sequence $S_{probe}$ belonging to identity $x_i$. Each sequence sample, $S = (S_{i, j}(1), S_{i, j}(2), \dots, S_{i, j}(T))$, is represented by a sequence of binary gait silhouettes $S_{i, j}(t)$ which depict the gait appearance in each frame.

\subsection{Model Overview}\label{sec:model_overview}
Our model is constructed by four main blocks as follows:
\begin{equation}\label{eq:model_preview}
    f = f_{PFE}\circ f_{RNN}\circ f_{CAP}\circ f_{CLS},
\end{equation}
where $F_{PFE}$ is a partial feature extraction of gait-maps constructed by multiple convolutional, pooling, and fully connected layers. $F_{RNN}$ is a bi-directional recurrent neural network which transforms the features extracted by the previous layer to a more discriminating manifold by exploring the spatial correlations between the horizontal strips in the feature maps. $f_{CAP}$ is a capsule attention layer which learns deeper part-whole relationships between the strips and then selectively assigns more weights to the more discriminative 
features and explains away misleading factors. The last layer,$F_{CLS}$, performs classification using a softmax activation function. Figure \ref{fig:2} represents the architecture of our model.

\subsection{Gait Partial Feature Extraction Block}\label{sec:primary_block}
The first block $F_{PFE}$ aims to extract discriminative feature bins where each bin corresponds to a partial spatial location on a spatio-temporal gait convolutional energy map. In this context, we have used Gaitset \cite{B1}, as the state-of-the-art deep partial feature extractor, to obtain multi-scale partial representations. Gaitset first uses eight 2D convolution and pooling layers to extract convolutional maps from individual silhouettes, $S_{i, j}(t)$ in a gait sequence. It then uses a set-pooling layer for summarizing the temporal dynamic behaviour of the gait by accumulating convolutional information temporally. This operation can be considered as an efficient temporal compression, applied to variable-length sequences of feature maps, thus making the resulting single map invariant to the number of frames for each sequence. Next, Gaitset partitions the gait map and represents each partial spatial location as a horizontal feature bin. This enables the model to learn about the salient and non-salient body parts, consequently making it less sensitive to missing key parts and local appearance variations. Finally, the split  bins at different scales, i.e., 1, 2, 4, 8, and 16, are mapped to a discriminative feature space using fully connected layers \cite{R9}, forming the gait partial feature representations.

\subsection{Recurrent Learning Block}\label{sec:recurrent_block}
Due to the existence of different body types and viewing angles, the saliency of feature bins varies from sequence to sequence. In another word, the first block of our model, Gaitset, is not invariant to spatial translations. To this end, we applied a BGRU \cite{GRU1} network that has been proven to be effective for many learning tasks from sequential data. Our BGRU network recurrently learns the correlations and co-occurrences of the patterns among the feature bins in forward and backward directions for each specific input representation. The output at each step is represented as a function of the current bin and the information preserved from both upper and lower feature bins. The advantage of using a bi-directional recurrent layer over a fully connected layer is two-fold. First, although a fully connected layer with enough capacity is capable of learning the correlation among feature bins, it would be likely to suffer from not being translation invariant, as opposed to the recurrent neural networks which are invariant to the (in this case) spatial translation. Second, a fully connected layer large enough to capture the discriminative correlations would require a large number of trainable parameters making it prone to overfitting especially for our scenario where we do not have a very large number of training samples across different viewing angles.

\subsection{Capsule Learning Block}\label{sec:attention_block}
Capsule network can decompose the input features into different hierarchical sub-parts to then develops a relationship between them. Unlike the standard pooling layers that lose positional attributes, such as location, rotation, and scale, capsule network can preserve these attributes. In this context, we propose to treat recurrently learned partial features as capsules to learn coupling weights between the underlying and output capsules. It exploits the hierarchical part-whole relations between the input partial feature representations while preserving positional attributes which helps the model to generalize better to novel viewpoints as one of the main challenges in the task of gait recognition. Capsule networks generally need fewer feature detectors and less training data due to their ability to deal with the viewpoint variation.

A capsule network is also capable of learning feature importance for determining the probe identity as it can selectively assign more weights to the more relevant features while ignoring the spurious dimensions \cite{Patrick}. One popular option compared to capsule network is using traditional attention layer where bi-directional recurrent hidden states are weighted by the learnt weights \cite{ATT2,ICASSP, joint}. However, to capture deeper spatial correlations among bins, multiple layers of attention are required. This results in a high expansion of model parameters making it prone to over-fitting especially when the number of training samples is limited. To this end, the capsule network also acts as an attention mechanism in our scenario, thus making our model more robust to different walking and appearance conditions.

The original capsule network proposed in \cite{caps} contains two main blocks, i.e., primary capsule and high-level capsule. Primary capsule block aims to encode spatial information using convolutional, reshaping and squashing layers, while high-level capsule block can then learn deeper part-whole relationships between hierarchical sub-parts. In this block we removed the convolutation layer from the primary capsule layer and only applied reshaping and squashing function to the outputs of the previous layer. The reason behind this selection is that the first block of our network, i.e., partial feature extraction, already encodes the spatial information through several convolutional layers, so we believe that no more convolutation learning is needed in this stage. Nevertheless, we will investigate the effects of using convolutational layer in the primary capsule layer on the gait recognition performance in Section 5.2.

First, the output generated by bi-directional recurrent layer is reshaped to a $C \times D_{1}$ tensor where $C$ is the number of primary capsules and $D_1$ is their dimension. This is followed by a non-linear squashing function which maps the length of each capsule to a number between 0 and 1. In our work we used the original version of squashing function as follows:
\begin{equation}\label{eq:squash_function}
    \mathbf{v}_{j}=\frac{\left\|\mathbf{s}_{j}\right\|^{2}}{1+\left\|\mathbf{s}_{j}\right\|^{2}} \frac{\mathbf{s}_{j}}{\left\|\mathbf{s}_{j}\right\|},
\end{equation}
where $v_j$ is the $j^{th}$ vector output of the primary capsule layer and $s_j$ is the $j^{th}$ capsule input formed by reshaping the output of recurrent learning block.
The second capsule layer is constructed by $D_2$  dimensional digit capsules where each capsule is computed by a routing process similar to originally proposed in \cite{caps}. Each unsquashed digit capsule is determined as a weighted sum of prediction vectors $\hat{\mathbf{u}}_{j | i}$ as follows:
\begin{equation}\label{eq:capsul_output}
    \mathbf{s}_{j}=\sum_{i} c_{i j} \hat{\mathbf{u}}_{j | i},
\end{equation}
where $c_{i j}s$ represents the coupling coefficients which are computed by the dynamic routing process. 
The prediction vectors $\hat{\mathbf{u}}_{j | i}$ are computed as a transformed version of capsule vectors in the previous layer as follows:
\begin{equation}\label{eq:prediction_vector}
    \hat{\mathbf{u}}_{j | i}=\mathbf{W}_{i j} \mathbf{u}_{i}
\end{equation}
where $\mathbf{W}_{i j}$ is a trainable weight matrix which encode the part-whole spatial relationships establishing an intrinsic capacity for predicting novel viewpoints. Thus, Eq. \ref{eq:capsul_output} can be considered as an attention layer over the transformed capsules $\hat{\mathbf{u}}_{j | i}$ where coupling coefficients are playing the role of attention weights.

\section{Experiments}\label{sec:experiments}
\subsection{Datasets}\label{sec:datasets}
Our experiments are conducted on two popular and large-scale gait datasets using four different test protocols.

\begin{figure*}[!t]
    \begin{center}
    \includegraphics[width=1\linewidth]{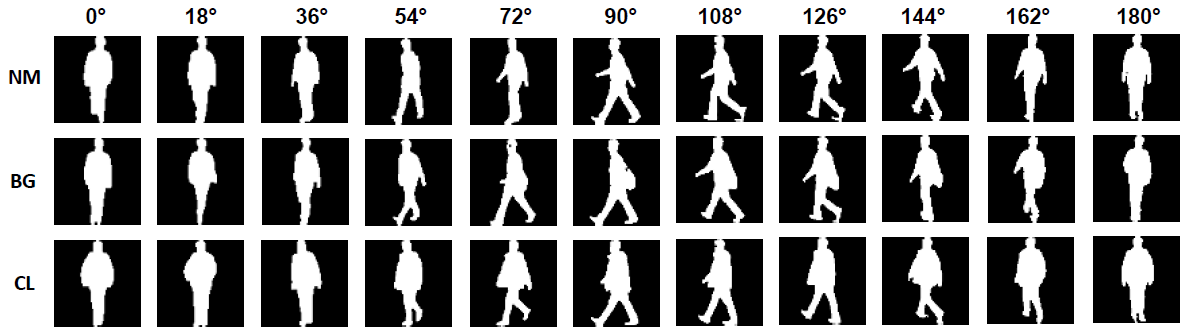} 
    \end{center}
\caption{Example cropped gait silhouettes from one subject in CASIA-B dataset \cite{CASIA}, captured in 11 evenly spaced angles for three different walking conditions.}
\label{fig:3}
\end{figure*}

{CASIA-B} \cite{CASIA} is a large and well-known gait dataset. It contains silhouette gait sequence data from 124 subjects in 3 different walking conditions and 11 different view angles. The walking conditions consist of normal walking (6 sequences per subject per view angle), walking with a bag (2 sequences per subject per view angle), and walking while wearing a coat (2 sequences per subject per view angle). The dataset was recorded in 11 evenly spaced angles over $0^\circ$ to $180^\circ$ ($0^\circ$, $18^\circ$, $36^\circ$, \dots, $180^\circ$), as shown in Figure \ref{fig:3}. In total, this dataset includes 13,750 gait sequences. 

\begin{figure*}[!t]
    \begin{center}
    \includegraphics[width=1\linewidth]{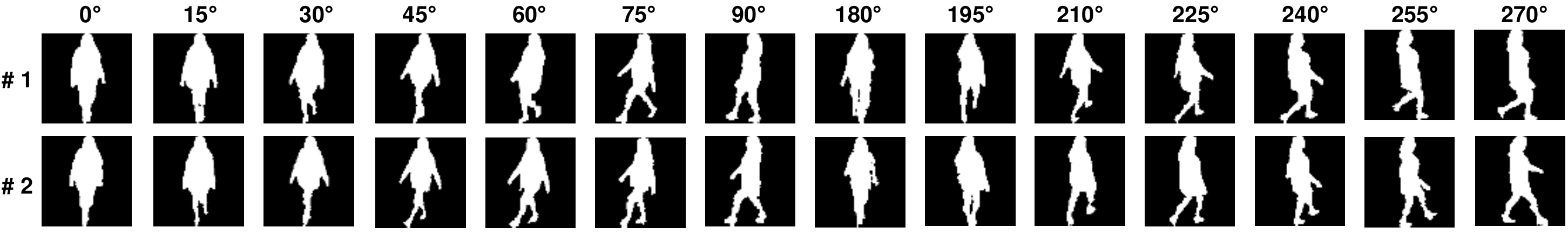}    
    \end{center}
\caption{Example cropped gait silhouettes from one subject in OU-MVLP dataset \cite{MVLP}, captured in 14 different view angles in two sessions.}
\label{fig:4}
\end{figure*}

{OU-MVLP} \cite{MVLP} is by far the largest available gait recognition dataset, recorded from 10,307 subjects (almost equally distributed in gender) in 14 different view angles \{$0^\circ$, $15^\circ$, \dots, $90^\circ$\} and \{$180^\circ$, $195^\circ$, \dots, $270^\circ$\}. There are 2 sequences per subject, as shown in Figure \ref{fig:4}. In total, the dataset includes 267,388 gait sequences.

\subsection{Implementation and Training Details}\label{sec:implementation}
We first calculate the bounding boxes of the gait silhouettes and then align and crop the gait silhouettes out of the frames. We finally resize all the cropped silhouettes into 64 $\times$ 64 pixels to feed our deep model. 

As presented in Section 3, our deep network includes four blocks, i.e., partial feature extraction, recurrent learning, capsule learning, and classification. We empirically tune the hyperparameter values for training our network in order to achieve the best performance, as presented in Table \ref{tab2}. This table presents the optimal hyperparameter values for each block of the network as well as for the whole deep network. It is worth noting that we use the pre-trained model available in \cite{Git} for extraction of Gaitset features \cite{B1}. Our entire architecture has been implemented using TensorFlow with Keras backend and is trained using four Nvidia RTX 2080 Ti GPUs.

\begin{table}
  \centering
  \setlength\tabcolsep{3pt}
\footnotesize
    \caption{Network hyperparameters.}
    \begin{tabular}{l|l|l}
    \hline
    \textbf {Block}& \textbf{Parameter}& \textbf{Value}\\
    \hline\hline
     Partial feature extraction & Mini-batch & Yes \\
     & Learning rate & 0.0001 \\
     & Loss function & $BA_{+}$ triplet loss \\
     & Optimizer & Adam \\
     \hline
     Recurrent learning & Recurrent depth & 1 \\
      & Hidden layer Size & 256 (128$\times$ 2) \\
      & Dropout rate & 0.25 \\
     \hline
     Capsule learning & Input feature map size &  16$\times$ 16 \\
     & Number of capsules & 6 \\
     & Capsule size & 128 \\
     & Number of routings & 3 \\
     \hline
     Classification & Activation & Softmax \\
     \hline
     Network & Mini-batch size (CASIA-B) & 50 \\
     & Mini-batch size (OU-MVLP) & 300 \\
     & Loss function & Cross-entropy \\
    \hline
    \end{tabular}%
  \label{tab2}%
\end{table}%

\subsection{Benchmarking and Test Protocols}\label{sec:benchmarking}
The state-of-the-art deep gait recognition methods presented in Table \ref{tab1} are used for benchmarking. All these methods have reported their results using the same datasets and test protocols that we also strictly followed. For the CASIA-B dataset \cite{CASIA}, we form the training and testing sets, respectively with the data from the first 74 subjects and the remaining 50 subjects. Then, the first four normal walking sequences are considered as the reference set, and the remaining 2 normal walking, 2 walking with coat, and 2 walking with bag sequences, form the NM, CL, BG probe sets respectively. The OU-MVLP dataset \cite{MVLP} is divided into two different recording sessions, one for training ($5153$ subjects) and one for testing ($5154$ subjects). During the test stage, one sequence (index \#00) is used as probe and the other sequence (index \#01) is held in the gallery. The results for both datasets have been reported for all combinations of angles, except the identical ones.

\begin{table*}[!t]
\centering
\caption{CASIA-B Gait recognition results under normal (NM) walking conditions.}
\setlength\tabcolsep{6pt}
\begin{tabular}{ l l l | l l l l l l l l l l l | l}
\hline
\multicolumn{3}{ c |}{\textbf{Method}} & \multicolumn{12}{ c }{\textbf{View}}  \\ 
\hline
	\textbf{Name} & \textbf{Year} & \textbf{Venue} & \textbf{0$^{\circ}$} & \textbf{18$^{\circ}$} & \textbf{36$^{\circ}$} & \textbf{54$^{\circ}$} & \textbf{72$^{\circ}$} & \textbf{90$^{\circ}$} & \textbf{108$^{\circ}$} & \textbf{126$^{\circ}$} & \textbf{144$^{\circ}$} & \textbf{162$^{\circ}$} & \textbf{180$^{\circ}$} & \textbf{Mean} \\ \hline\hline
	CNN-3D ~\cite{B6}& 2017 & \textit{T-PAMI} & 87.1 & 93.2 & 97.0 & 94.6 & 90.2 & 88.3 & 91.1 & 93.8 & 96.5 & 96.0 & 85.7 & 92.1 \\ 
	CNN-Ens.~\cite{B6} & 2017 & \textit{T-PAMI} & 88.7 & 95.1 & 98.2 & 96.4 & 94.1 & 91.5 & 93.9 & 97.5 & 98.4 & 95.8 & 85.6 & 94.1 \\ 
	MGAN  ~\cite{B7}& 2019 & \textit{T-IFS} & - & - & - & 84.2 & - & 72.3 & - & 83.0 & - & - & - & 79.8 \\ 
	EV-Gait~\cite{B2} & 2019 & \textit{CVPR} & 77.3 & 89.3 & 94.0 & 91.8 & 92.3 & 96.2 & 91.8 & 91.8 & 91.4 & 87.8 & 85.7 & 89.9 \\ 
	Gait-Joint ~\cite{B10} & 2019 & \textit{PR} & 75.6 & 91.3 & 91.2 & 92.9 & 92.5 & 91.0 & 91.8 & 93.8 & 92.9 & 94.1 & 81.9 & 89.9\\
	GaitSet ~\cite{B1}& 2019 & \textit{AAAI} & 90.8 & 97.9 & \textbf{99.4} & 96.9 & 93.6 & 91.7 & 95.0 & 97.8 & \textbf{98.9} & \textbf{96.8} & 85.8 & 95.0 \\ 
	GaitNet-1~\cite{B3} & 2019 & \textit{CVPR} & 91.2 & - & - & 95.6 & - & 92.6 & - & 96.0 & - & - & - & 93.9 \\
	GaitNet-2 ~\cite{B4} & 2019 & \textit{arXiv} & \textbf{93.1} & 92.6 & 90.8 & 92.4 & 87.6 & \textbf{95.1} & 94.2 & 95.8 & 92.6 & 90.4 & \textbf{90.2} & 92.3 \\ 
	\textbf{Ours} & - & - & 91.8 & \textbf{98.3} & 99.0 & \textbf{98.0} & \textbf{94.1} & 92.8 & \textbf{96.3} & \textbf{98.1} & 98.4 & 96.2 & 89.2 & \textbf{95.7}  \\ 
\hline
\end{tabular}
\label{tab3}
\end{table*}

\begin{table*}[!t]
\centering
\caption{ CASIA-B Gait recognition results under carried bags (BG) conditions.}
\setlength\tabcolsep{6pt}
\begin{tabular}{ l l l | l l l l l l l l l l l | l}
\hline
\multicolumn{3}{ c |}{\textbf{Method}} & \multicolumn{12}{ c }{\textbf{View}}  \\
\hline
	\textbf{Name} & \textbf{Year} & \textbf{Venue} & \textbf{0$^{\circ}$} & \textbf{18$^{\circ}$} & \textbf{36$^{\circ}$} & \textbf{54$^{\circ}$} & \textbf{72$^{\circ}$} & \textbf{90$^{\circ}$} & \textbf{108$^{\circ}$} & \textbf{126$^{\circ}$} & \textbf{144$^{\circ}$} & \textbf{162$^{\circ}$} & \textbf{180$^{\circ}$} & \textbf{Mean} \\ \hline\hline
	CNN-LB ~\cite{B6}& 2017 & \textit{T-PAMI} & 64.2 & 80.6 & 82.7 & 76.9 & 64.8 & 63.1 & 68 & 76.9 & 82.2 & 75.4 & 61.3 & 72.4 \\ 
	MGAN ~\cite{B7}& 2019 & \textit{T-IFS} & 48.5 & 58.5 & 59.7 & 58 & 53.7 & 49.8 & 54 & 61.3 & 59.5 & 55.9 & 43.1 & 54.7 \\ 
	GaitSet ~\cite{B1}&2019  & \textit{AAAI} & 83.8 & 91.2 & 91.8 & 88.8 & 83.3 & 81.0 & 84.1 & 90.0 & 92.2 & \textbf{94.4} & 79 & 87.2 \\ 
	GaitNet-1 ~\cite{B3}&2019  & \textit{CVPR} & 83.0 & - & - & 86.0 & - & 74.8 & - & 85.8 & - & - & - & 82.6 \\ 
	GaitNet-2 ~\cite{B4}&2019  & \textit{arXiv} & \textbf{88.8} & 88.7 & 88.7 & \textbf{94.3} & 85.4 & \textbf{92.7} & \textbf{91.1} & 92.6 & 84.9 & 84.4 & \textbf{86.7} & 88.9 \\
	\textbf{Ours} & - & - & 87.3 & \textbf{93.7} & \textbf{94.8} & 93.1 & \textbf{88.1} & 84.5 &  88.8 & \textbf{93.5 } & \textbf{96.3 } & 93.3  & 83.9  & \textbf{90.7 } \\
\hline
\end{tabular}
\label{tab4}
\end{table*}

\begin{table*}[!t]
\centering
\caption{CASIA-B Gait recognition results under different clothing (CL) conditions.}
\setlength\tabcolsep{6pt}
\begin{tabular}{ l l l | l l l l l l l l l l l | l}
\hline
\multicolumn{3}{ c |}{\textbf{Method}} & \multicolumn{12}{ c }{\textbf{View}}  \\ 
\hline
	\textbf{Name} & \textbf{Year} & \textbf{Venue} & \textbf{0$^{\circ}$} & \textbf{18$^{\circ}$} & \textbf{36$^{\circ}$} & \textbf{54$^{\circ}$} & \textbf{72$^{\circ}$} & \textbf{90$^{\circ}$} & \textbf{108$^{\circ}$} & \textbf{126$^{\circ}$} & \textbf{144$^{\circ}$} & \textbf{162$^{\circ}$} & \textbf{180$^{\circ}$} & \textbf{Mean} \\ \hline\hline
	CNN-LB ~\cite{B6}& 2017 & \textit{T-PAMI} & 37.7 & 57.2 & 66.6 & 61.1 & 55.2 & 54.6 & 55.2 & 59.1 & 58.9 & 48.8 & 39.4 & 54.0 \\ 
	MGAN ~\cite{B7}& 2019 & \textit{T-IFS} & 23.1 & 34.5 & 36.3 & 33.3 & 32.9 & 32.7 & 34.2 & 37.6 & 33.7 & 26.7 & 21.0 & 31.5 \\ 
	GaitSet ~\cite{B1}& 2019 & \textit{AAAI} & 61.4 & 75.4 & 80.7 & 77.3 & 72.1 & 70.1 & \textbf{71.5} & 73.5 & 73.5 & 68.4 & 50 & 70.4\\ 
	GaitNet-1~\cite{B3} & 2019 & \textit{CVPR} & 42.1 & - & - & 70.7 & - & 70.6 & - & 69.4 & - & - & - & 63.2 \\ 
	GaitNet-2 ~\cite{B4}& 2019 & \textit{arXiv} & 50.1 & 60.7 & 72.4 & 72.1 & \textbf{74.6} & \textbf{78.4} & 70.3 & 68.2 & 53.5 & 44.1 & 40.8 & 62.3 \\ 
	\textbf{Ours} & - & - & \textbf{63.4} & \textbf{77.3} & \textbf{80.1} & \textbf{79.4} & 72.4 & 69.8 & 71.2 & \textbf{73.8} & \textbf{75.5} & \textbf{71.7} & \textbf{62.0} & \textbf{72.4}\\  
\hline
\end{tabular}
\label{tab5}
\end{table*}

\begin{table}[!t]
\centering
\setlength\tabcolsep{5pt}
\caption{OU-MVLP gait recognition results.}
\begin{tabular}{ l l l | l l l l | l }
\hline
\multicolumn{3}{ c |}{\textbf{Method}} & \multicolumn{5}{ c }{\textbf{View}}  \\ 
\hline
	\textbf{Name} & \textbf{Year} & \textbf{Venue} &  \textbf{0$^{\circ}$} & \textbf{30$^{\circ}$} & \textbf{60$^{\circ}$} & \textbf{90$^{\circ}$} & \textbf{Mean} \\ \hline\hline
	GEI NET ~\cite{B8}& 2016 & \textit{ICB} & 15.7 & 41 & 39.7 & 39.5 & 34.0 \\ 
	CNN-LB ~\cite{B6}& 2017 & \textit{T-PAMI} & 14.2 & 32.7 & 32.3 & 34.6 & 28.5 \\ 
	DigGAN ~\cite{B9} & 2018 & \textit{arXiv} & 30.8 & 43.6 & 41.3 & 42.5 & 39.6 \\ 
	Gait-Set ~\cite{B1}& 2019 & \textit{AAAI} & 77.7 & 86.9 & 85.3 & 83.5 & 83.4 \\ 
	\textbf{Ours}& - & - & \textbf{78.3} & \textbf{88.8} & \textbf{85.7} & \textbf{85.1} & \textbf{84.5} \\
\hline
\end{tabular}
\label{tab6}
\end{table}

\section{Results}\label{sec:results}

\subsection{Performance}\label{sec:performance}
Tables \ref{tab3}, \ref{tab4}, and \ref{tab5} show the rank-1 recognition results obtained by our and other state-of-the-art methods on the CASIA-B dataset, for three different walking conditions including normal walking (NM), walking while carrying a bag (BG), and walking while wearing a coat (CL), respectively. In addition, Table \ref{tab6} compares the rank-1 recognition results obtained by our model and other state-of-the-art methods on the OU-MVLP dataset.

\subsubsection{Gait recognition under different walking conditions} As shown in Tables \ref{tab3}, \ref{tab4}, and \ref{tab5}, the average gait recognition values obtained by our method are consistently better than other state-of-the-art methods for all the test protocols considered. We achieve average performance gains of 0.7\%, 1.8\%, 2.0\%, when compared to the best performing benchmarking methods \cite{B1}\cite{B4} on CASIA-B dataset for NM, BG, and CL walking conditions respectively. Compared to the best performing benchmarking method \cite{B1} on the OU-MVLP dataset, we achieve a performance gain of 1.1\%. The superior performance of our method in more evident when facing challenging appearance conditions, i.e., walking while carrying a bag (Table \ref{tab4}), and wearing a coat (Table \ref{tab5}). This is due to the adaptation of our capsule network which is able to learn deeper part-whole relationships and selectively focus on the most informative partial gait representations, thus making our model more robust to these challenging variations.

\subsubsection{Gait recognition under different viewing angles} The results in Tables 3 through 6 show that our method reaches the best results for most of the viewing angles. As the results show, our model obtained less impressive results when facing the extreme viewing angles, including those that are parallel to the walking direction, e.g, 0$^{\circ}$ and 180$^{\circ}$, and those that are vertical to the walking direction, e.g., 90$^{\circ}$. These extreme viewing angles cause for some useful gait information to be missed such as stride and body swing, naturally reducing the recognition performance. Nevertheless, the performance of our network presented in Tables 3 through 6 is better than the state-of-the-art methods in most of these difficult angles, showing that our model is able to learn more robust features for extreme viewing angle changes. Our method achieves very impressive results for the gait data captured from intermediate angles, where useful gait information such as body shape and walking postures is visible. Finally, the results show that our model also performs very well when facing intermediate and extreme angles for training and testing and vice versa.

\subsection{Feature Space Exploration}

\begin{figure}[!t]
\centering
\includegraphics[width=1\columnwidth]{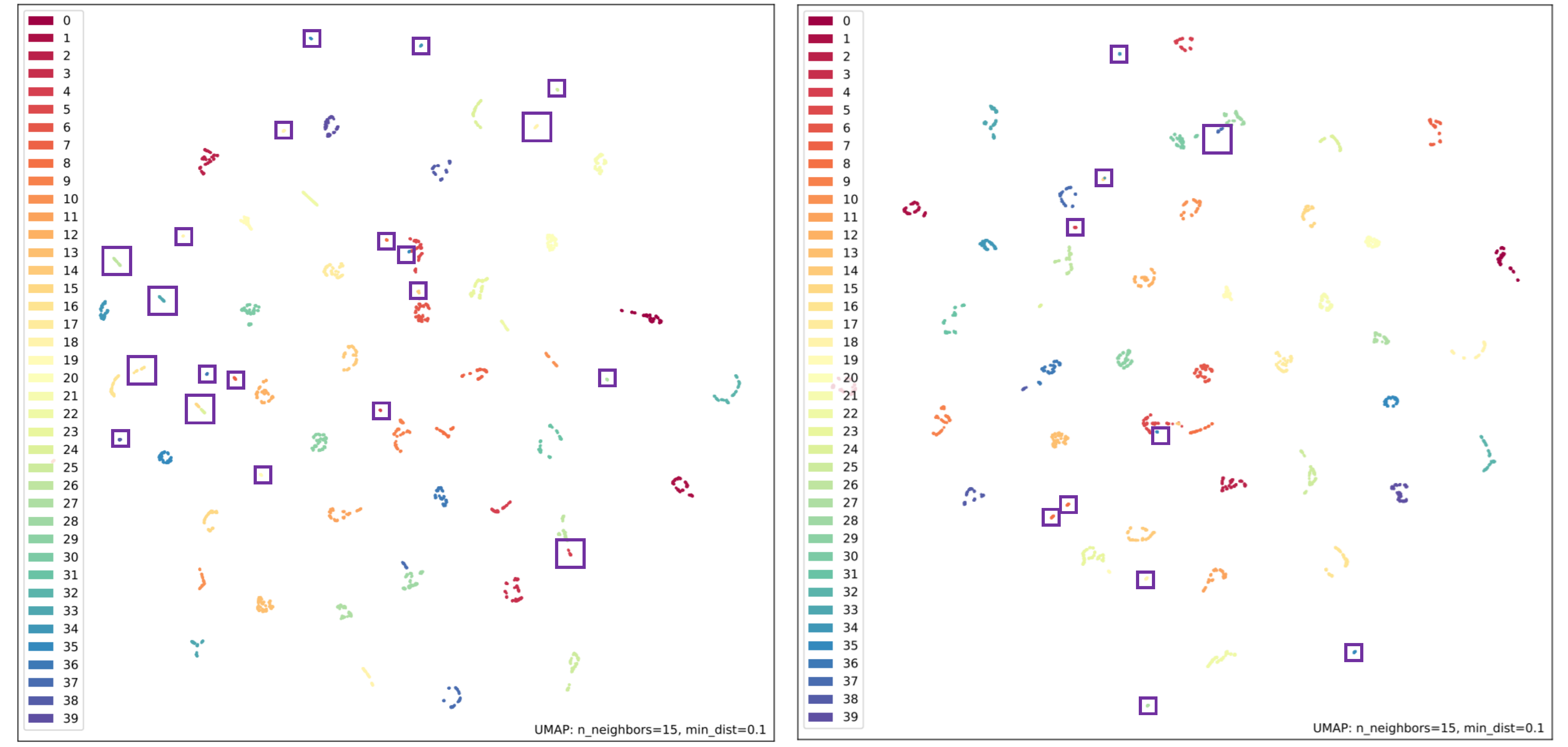}
\caption{UMAP visualization of the feature spaces produced by (left) global feature representation; and (right) our proposed solution.}
\label{fig:UMAP}
\end{figure}

UMAP visualization \cite{umap} has been used to demonstrate the discrimination ability of our proposed solution. This visualization shows the feature spaces produced by our proposed solution and the global feature representation approach when dealing with gait data as a whole, in Figure \ref{fig:UMAP}. This visualisation is performed for the first 40 subjects available in CASIA dataset. A better representation should create less data points distributed far from their respective cluster's centroid. These data points that may lead to misclassification are denoted by purple boxes for the baseline and our proposed solutions, respectively in Figure \ref{fig:UMAP}-left and Figure \ref{fig:UMAP}-right. As the results show, our solution creates denser clusters, thus the subjects are more easily separable in our feature spaces, compared to global feature representation.

\subsection{Ablation Study}\label{sec:ablation_study}
We have performed an ablation study on CASIA-B dataset \cite{CASIA} using NM, CL, BG test protocols to investigate the effects of each of our proposed network's components towards the final recognition performance. Table \ref{tab7} presents the performance of our model when removing individual blocks of the network. It should be noted that we always keep the Gaitset \cite{B1} block in order to extract multi-scale partial feature representations, where its individual results are presented in the first row of Table \ref{tab7}. We then remove the layers from our network systematically and one component at a time. The performance of the complete model is also presented in the last row of Table \ref{tab7}. The following subsections describe each ablation test.

\subsubsection{Impact of bi-directional recurrent learning} We start by removing the second GRU layer, thus turning BGRU to a uni-directional GRU. This means that we would only exploit the forward relationships within gait sequences for recurrent learning. The results in the second row of Table \ref{tab7} show that the performance decreased by 0.8\%, thus revealing the added value of exploiting both forward and backward relationships within gait sequences using BGRU.

\subsubsection{Impact of recurrent learning} Next, we remove the whole recurrent learning block, i.e., BGRU, thus feeding the capsule network with partial features obtained by Gaitset. The results presented in the third row of Table \ref{tab7} clearly show the degradation of the results from 86.3\% to 85.3\%, thus demonstrating the benefit of the recurrent learning block in learning the relations between multi-scale partial gait representations.  

\subsubsection{Impact of capsule learning} We finally remove the capsule layer, thus preventing the part-whole relationships to be learned. This implies that the recurrently learned partial features maintain the same importance towards the recognition process and their hierarchical relations are not exploited. The results in the fourth row of Table \ref{tab7} show that the performance drops by 1.2\% when removing this block. The added benefit of the capsule layer is especially more evident when facing challenging appearance conditions, namely walking while carrying a bag or wearing a coat.

\begin{table}[!t]
\centering
\footnotesize
\centering
\caption{Ablation study on the CASIA-B dataset.}
\setlength
\tabcolsep{4pt}
\begin{tabular}{ l| l| l| l| l| l| l| l}
\hline
 \multicolumn{1}{ c |}{\textbf{GaitSet}}&\multicolumn{1}{ c |}{\textbf{GRU 1}}& \multicolumn{1}{ c |}{\textbf{GRU 2}}& \multicolumn{1}{ c| }{\textbf{ Capsule}}& \multicolumn{4}{ c }{\textbf{Protocol}}  \\ 
\hline
    & & & & NM & BG & CL & Mean\\ \hline\hline
    \cmark & \xmark & \xmark & \xmark & 95.0 & 87.2 & 70.4 & 84.2 \\
    \cmark & \cmark & \xmark & \cmark &  95.2& 89.6 & 71.7 & 85.5\\
    \cmark & \xmark & \xmark & \cmark & 95.3 & 89.9 & 70.7 & 85.3\\
    \cmark & \cmark & \cmark & \xmark & 95.0 & 90.0 & 70.4 & 85.1\\
    \cmark & \cmark & \cmark & \cmark & 95.7 & 90.7 & 72.4 & 86.3\\
    \hline
\end{tabular}
\label{tab7}
\end{table}

\subsubsection{Impact of convolutional layer in primary capsule block}
As discussed in Section 3.5, the original capsule network proposed in \cite{caps} contains two main blocks, i.e., primary capsule and high-level capsule. In our proposed architecture, we did not utilize any primary capsule convolutional layers due to the fact that the first block of our network, i.e., partial feature extraction, already encodes the spatial information through several convolutional layers. Here, we also investigate the effects of adding the primary capsule convolutional layer to the architecture. In this context, the output generated by each cell in the bi-directional recurrent layer with the size of 256 is first reshaped to a 2D $16 \times 16$ feature map. Then, all these feature maps are stacked along the third dimension to form the local features that are used as input to the convolutional layer. This modification not only considerably increased the computational complexity of our network, but also decreased the average gait recognition performance from 86.3\% to 84.2\%. This phenomena is due to the fact that applying convolutional kernels in the primary capsule block results in losing some information available in the border of the reshaped feature maps, which is potentially the reason for the lower performance of the recognition system when considering the convolutional layer in primary capsule block.

\section{Conclusion}
We proposed a novel deep model for transferring multi-scale partial features into a more discriminative set of representations that is more robust to challenging gait recognition variations such as viewing angle, clothing, and an obstructive item namely a bag. Our proposed network recurrently learns the relations between the multi-scale partial features that are previously extracted using a convolution-based feature extractor. It then uses a capsule network to learn deeper part-whole relationships and to act as an attention mechanism for making our model more robust to different viewing and appearance conditions. We have extensively tested our method on CASIA-B and OU-MVLP datasets. Our proposed method outperformed the benchmarking state-of-the-art methods using different and challenging test protocols.

\section{Acknowledgment}
The authors would like to thank the BMO Bank of Montreal and Mitacs for funding this research.

\bibliographystyle{ieeetran}
\bibliography{egbib}

\begin{thebibliography}{10}
\providecommand{\url}[1]{#1}
\csname url@samestyle\endcsname
\providecommand{\newblock}{\relax}
\providecommand{\bibinfo}[2]{#2}
\providecommand{\BIBentrySTDinterwordspacing}{\spaceskip=0pt\relax}
\providecommand{\BIBentryALTinterwordstretchfactor}{4}
\providecommand{\BIBentryALTinterwordspacing}{\spaceskip=\fontdimen2\font plus
\BIBentryALTinterwordstretchfactor\fontdimen3\font minus
  \fontdimen4\font\relax}
\providecommand{\BIBforeignlanguage}[2]{{%
\expandafter\ifx\csname l@#1\endcsname\relax
\typeout{** WARNING: IEEEtran.bst: No hyphenation pattern has been}%
\typeout{** loaded for the language `#1'. Using the pattern for}%
\typeout{** the default language instead.}%
\else
\language=\csname l@#1\endcsname
\fi
#2}}
\providecommand{\BIBdecl}{\relax}
\BIBdecl

\bibitem{R4}
C.~Wan, L.~Wang, and V.~Phoha, ``A survey on gait recognition,'' \emph{ACM
  Computing Surveys}, vol.~51, no.~5, pp. 89--124, August 2018.

\bibitem{nambiar}
A.~Nambiar, A.~Bernardino, and J.~C. Nascimento, ``Gait-based person
  re-identification: A survey,'' \emph{ACM Computing Surveys}, vol.~52, no.~2,
  pp. 1--34, April 2019.

\bibitem{R2}
I.~Rida, N.~Almaadeed, and S.~A. 1, ``Robust gait recognition: a comprehensive
  survey,'' \emph{IET Biometrics}, vol.~8, no.~1, pp. 14--28, January 2019.

\bibitem{wear}
M.~D. Marsico and A.~Mecca, ``A survey on gait recognition via wearable
  sensors,'' \emph{ACM Computing Surveys}, vol.~52, no.~4, pp. 1--39, September
  2019.

\bibitem{B6}
Z.~{Wu}, Y.~{Huang}, L.~{Wang}, X.~{Wang}, and T.~{Tan}, ``A comprehensive
  study on cross-view gait based human identification with deep {CNNs},''
  \emph{IEEE Transactions on Pattern Analysis and Machine Intelligence},
  vol.~39, no.~2, February 2017.

\bibitem{ICPR1}
{Shiqi Yu}, {Daoliang Tan}, and {Tieniu Tan}, ``A framework for evaluating the
  effect of view angle, clothing and carrying condition on gait recognition,''
  in \emph{International Conference on Pattern Recognition}, Hong Kong, China,
  September 2006.

\bibitem{TanmayThesis}
T.~Verlekar, ``Gait analysis in unconstrained environments,'' Ph.D.
  dissertation, Electrical and Computer Engineering, Instituto Superior
  Técnico, University of Lisbon, Lisbon, Portugal, June 2020.

\bibitem{view}
T.~Verlekar, P.~Correia, and L.~Soares, ``View-invariant gait recognition
  system using a gait energy image decomposition method,'' \emph{IET
  Biometrics}, vol.~6, no.~4, pp. 299--306, July 2017.

\bibitem{R6}
X.~{Li}, Y.~{Makihara}, C.~{Xu}, Y.~{Yagi}, and M.~{Ren}, ``Joint intensity
  transformer network for gait recognition robust against clothing and carrying
  status,'' \emph{IEEE Transactions on Information Forensics and Security},
  vol.~14, no.~12, pp. 3102--3115, December 2019.

\bibitem{B1}
H.~Chao, Y.~He, J.~Zhang, and J.~Feng, ``Gaitset: Regarding gait as a set for
  cross-view gait recognition,'' in \emph{AAAI Conference on Artificial
  Intelligence}, Honolulu, HW, USA, February 2019.

\bibitem{GRU1}
K.~Cho, B.~V. Merri{\"e}nboer, D.~Bahdanau, and Y.~Bengio, ``On the properties
  of neural machine translation: Encoder-decoder approaches,''
  \emph{arXiv:1409.1259}, October 2014.

\bibitem{B8}
K.~{Shiraga}, Y.~{Makihara}, D.~{Muramatsu}, T.~{Echigo}, and Y.~{Yagi},
  ``{GEINet}: View-invariant gait recognition using a convolutional neural
  network,'' in \emph{International Conference on Biometrics}, Halmstad,
  Sweden, June 2016.

\bibitem{B9}
B.~Hu, Y.~Gao, Y.~Guan, Y.~Long, N.~Lane, and T.~Ploetz, ``Robust cross-view
  gait identification with evidence: A discriminant gait {GAN} {(DiGGAN)}
  approach on 10000 people,'' \emph{arXiv:1811.10493}, 2018.

\bibitem{B7}
Y.~{He}, J.~{Zhang}, H.~{Shan}, and L.~{Wang}, ``Multi-task {GANs} for
  view-specific feature learning in gait recognition,'' \emph{IEEE Transactions
  on Information Forensics and Security}, vol.~14, no.~1, pp. 102--113, January
  2019.

\bibitem{B2}
Y.~Wang, B.~Du, Y.~Shen, K.~Wu, G.~Zhao, J.~Sun, and H.~Wen, ``{EV-Gait}:
  Event-based robust gait recognition using dynamic vision sensors,'' in
  \emph{Computer Vision and Pattern Recognition}, Long Beach, CA, USA, June
  2019.

\bibitem{B10}
S.~Yu, R.~Liao, W.~An, H.~Chen, E.~García, Y.~Huang, and N.~Poh, ``Gaitnet: An
  end-to-end network for gait based human identification,'' \emph{Pattern
  Recognition}, vol.~96, no.~1, pp. 1--11, December 2019.

\bibitem{B3}
Z.~Zhang, L.~Tran, X.~Yin, Y.~Atoum, X.~Liu, J.~Wan, and N.~Wang, ``Gait
  recognition via disentangled representation learning,'' in \emph{Computer
  Vision and Pattern Recognition}, Long Beach, CA, USA, June 2019.

\bibitem{B4}
Z.~Zhang, L.~Tran, F.~Liu, and X.~Liu, ``On learning disentangled
  representations for gait recognition,'' \emph{arXiv preprint
  arXiv:1909.03051}, September 2019.

\bibitem{R9}
Y.~Fu, Y.~Wei, Y.~Zhou, H.~Shi, G.~Huang, X.~Wang, Z.~Yao, and T.~Huang,
  ``Horizontal pyramid matching for person re-identification,'' in \emph{AAAI
  Conference on Artificial Intelligence}, Honolulu, HW, USA, February 2019.

\bibitem{part}
R.~Imad, ``Towards human body-part learning for model-free gait recognition,''
  \emph{arXiv:1904.01620}, April 2019.

\bibitem{caps}
S.~Sabour, N.~Frosst, and G.~E. Hinton, ``Dynamic routing between capsules,''
  in \emph{Advances in neural information processing systems (NIPS)}, Long
  Beach, CA, USA, December 2017, pp. 3856--3866.

\bibitem{GRUCaps1}
S.~Chen, B.~Zheng, and T.~Hao, ``Capsule-based bidirectional gated recurrent
  unit networks for question target classification,'' in \emph{China Conference
  on Information Retrieval}, Guilin, China, September 2018.

\bibitem{GRUCaps2}
Y.~Du, X.~Zhao, M.~He, and W.~Guo, ``A novel capsule based hybrid neural
  network for sentiment classification,'' \emph{IEEE Access}, vol.~7, pp.
  39\,321--39\,328, March 2019.

\bibitem{GRUCaps3}
P.~Rathnayaka, S.~Abeysinghe, C.~Samarajeewa, I.~Manchanayake, and M.~Walpola,
  ``Gated recurrent neural network and capsule network based approach for
  implicit emotion detection,'' \emph{arXiv preprint arXiv:1809.01452},
  September 2018.

\bibitem{Model}
G.~{Ariyanto} and M.~{Nixon}, ``Model-based {3D} gait biometrics,'' in
  \emph{International Joint Conference on Biometrics}, Crete, Greece, September
  2010.

\bibitem{R7}
T.~Verlekar, L.~Soares, and P.~Correia, ``Gait recognition in the wild using
  shadow silhouettes,'' \emph{Image and Vision Computing}, vol.~76, no.~1, pp.
  1 -- 13, August 2018.

\bibitem{ICPR2}
{Yang Feng}, {Yuncheng Li}, and {Jiebo Luo}, ``Learning effective gait features
  using {LSTM},'' in \emph{International Conference on Pattern Recognition},
  Cancun, Mexico, April 2016.

\bibitem{GEI1}
C.~{Wang}, J.~{Zhang}, p.~{Xiaoru}, and L.~{Wang}, ``{Chrono-Gait} image: A
  novel temporal template for gait recognition,'' in \emph{European Conference
  on Computer Vision}, Washington, DC, USA, December 2011.

\bibitem{Patrick}
G.~Zhang and A.~Etemad, ``Capsule attention for multimodal {EEG} and {EOG}
  spatiotemporal representation learning with application to driver vigilance
  estimation,'' \emph{arXiv preprint:1912.07812}, January 2020.

\bibitem{ATT2}
G.~{Zhang}, V.~{Davoodnia}, A.~{Sepas-Moghaddam}, Y.~{Zhang}, and A.~{Etemad},
  ``Classification of hand movements from {EEG} using a deep attention-based
  {LSTM} network,'' \emph{IEEE Sensors Journal}, vol.~20, no.~6, March 2020.

\bibitem{ICASSP}
A.~Sepas-Moghaddam, A.~Etemad, F.~Pereira, and P.~L. Correia, ``Facial emotion
  recognition using light field images with deep attention-based bidirectional
  {LSTM},'' in \emph{IEEE conference on Acoustics, Speech, and Signal
  Processing}, Barcelona, Spain, May 2020.

\bibitem{joint}
A.~Sepas-Moghaddam, A.~Etemad, F.~Pereira, and P.~Correia, ``Long short-term
  memory with gate and state level fusion for light field-based face
  recognition,'' \emph{arXiv 1905.04421}, May 2020.

\bibitem{CASIA}
S.~Yu, D.~Tan, and T.~Tan, ``A framework for evaluating the effect of view
  angle, clothing and carrying condition on gait recognition,'' in
  \emph{International Conference on Pattern Recognition}, Hong Kong, China,
  August 2006.

\bibitem{MVLP}
N.~Takemura, Y.~Makihara, D.~Muramatsu, T.~Echigo, and Y.~Yagi, ``Multi-view
  large population gait dataset and its performance evaluation for cross-view
  gait recognition,'' \emph{Pattern Recognition}, vol.~72, no.~1, pp. 123--143,
  December 2017.

\bibitem{Git}
\BIBentryALTinterwordspacing
H.~Chao, Y.~He, J.~Zhang, and J.~Feng. (2019) A flexible, effective and fast
  cross-view gait recognition network. [Online]. Available:
  \url{hhttps://github.com/AbnerHqC/GaitSet}
\BIBentrySTDinterwordspacing

\bibitem{umap}
L.~McInnes, J.~Healy, and J.~Melville, ``Umap: Uniform manifold approximation
  and projection for dimension reduction,'' \emph{arXiv preprint
  arXiv:1802.03426}, December 2018.

\end{thebibliography}

\end{document}